\documentclass[11pt]{article}

\usepackage{acl}  

\usepackage{times}
\usepackage{latexsym}

\usepackage[T1]{fontenc}

\usepackage[utf8]{inputenc}

\usepackage{microtype}
\usepackage{multirow}
\usepackage{booktabs}
\usepackage{graphicx}
\usepackage{tabularx}
\usepackage{amsmath}
\usepackage{listings}
\usepackage{csquotes}

\usepackage{array}
\newcolumntype{L}[1]{>{\raggedright\let\newline\\\arraybackslash\hspace{0pt}}m{#1}}
\newcolumntype{C}[1]{>{\centering\let\newline\\\arraybackslash\hspace{0pt}}m{#1}}
\newcolumntype{R}[1]{>{\raggedleft\let\newline\\\arraybackslash\hspace{0pt}}m{#1}}

\title{A Usage-centric Take on Intent Understanding in E-Commerce}

\author{Wendi Zhou$^{1\dagger}$ \quad Tianyi Li$^{1}$ \quad Pavlos Vougiouklis$^{2}$  
        \quad {\bf Mark Steedman}$^{1}$ \quad {\bf Jeff Z. Pan}$^{1,2}$\footnotemark{*}  \\
        $^{1}$University of Edinburgh \quad $^{2}$Huawei Technologies,
Edinburgh RC, CSI\\
\texttt{\{}{\href{mailto:s2236454@ed.ac.uk}{\texttt{s2236454}}}\texttt{,}
{\href{mailto:tianyi.li@ed.ac.uk}{\texttt{tianyi.li}}}\texttt{,}
{\href{mailto:m.steedman@ed.ac.uk}{\texttt{m.steedman}}}\texttt{\}@ed.ac.uk}\\
\texttt{\{}{\href{mailto:pavlos.vougiouklis@huawei.com}{\texttt{pavlos.vougiouklis}}}\texttt{\}@huawei.com}\\
 \href{http://knowledge-representation.org/j.z.pan/}{http://knowledge-representation.org/j.z.pan/}}

\begin{document}
\maketitle
\begin{abstract}
Identifying and understanding user intents is a pivotal task for E-Commerce. Despite its 
essential role in product recommendation and business user profiling analysis, 
intent understanding has not been consistently defined or accurately benchmarked.
In this paper, we focus on predicative user intents as ``how a customer uses a product'', and pose intent understanding as a natural language reasoning task, independent of product ontologies. 
We identify two weaknesses of FolkScope, the SOTA E-Commerce Intent Knowledge Graph: \textit{category-rigidity} and \textit{property-ambiguity}.
They limit its ability to 
strongly align user intents with products having the most desirable property, and to recommend useful products across diverse categories. Following these observations, we introduce a Product Recovery Benchmark featuring a novel evaluation framework and an example dataset. We further validate the above FolkScope weaknesses 
on this benchmark. Our code and dataset are available at \href{https://github.com/stayones/Usgae-Centric-Intent-Understanding}{https://github.com/stayones/Usgae-Centric-Intent-Understanding}.

\end{abstract}

\section{Introduction}
\renewcommand{\thefootnote}{\fnsymbol{footnote}}
\setcounter{footnote}{1}
\footnotetext{Contact Author}
\renewcommand{\thefootnote}{\arabic{footnote}}

\renewcommand{\thefootnote}{\fnsymbol{footnote}}
\setcounter{footnote}{2}
\footnotetext{Work done while at Huawei Edinburgh Research Centre.}
\renewcommand{\thefootnote}{\arabic{footnote}}
\setcounter{footnote}{0}

User intents are a crucial source of information for E-Commerce \cite{DWLZD+2023,EOJP2023,10.1145/2872427.2874810,10.1145/3488560.3498517}. Intents reveal users' motivation in E-Commerce interactions: suppose a user plans to go for \textcolor{olive}{outdoor barbecue}, 
their intent may not refer only to barbeque smoker grills but also to other products that can be useful, such as disposable cutlery or plates.
In these cases, traditional product recommendation approaches would fail to handle these queries or to remind customers of the products they may need but have forgotten.
\begin{figure}
    \centering
    \includegraphics[width=0.95\columnwidth]{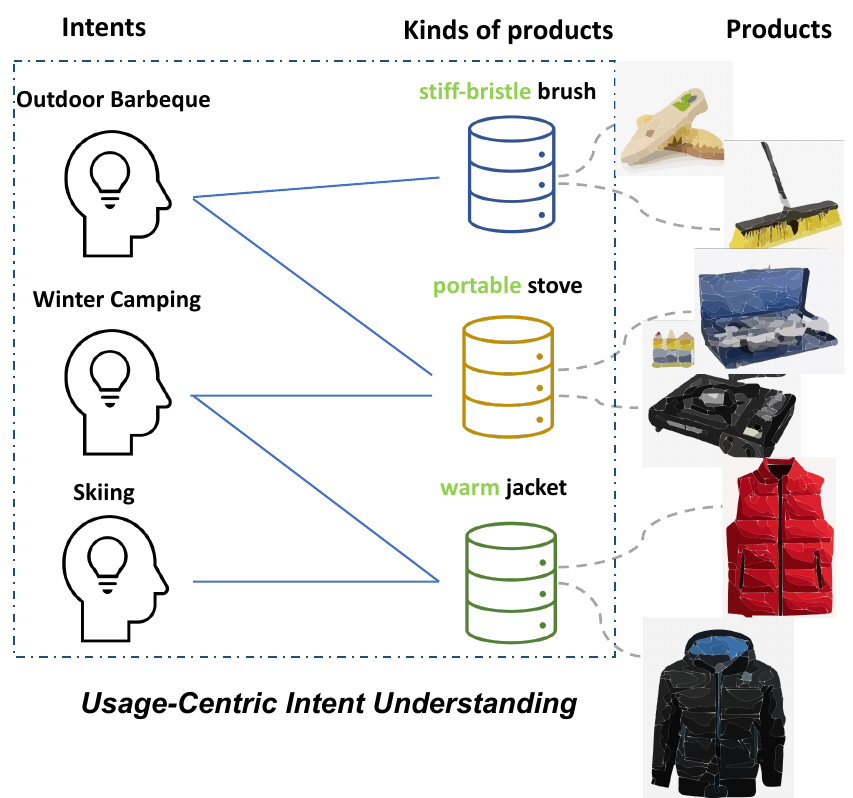}
    \caption{A graphic illustration of the usage-centric paradigm of intent understanding.}
    \label{fig:general_describe}
\end{figure}
\textit{Intent Understanding} offers 
great benefits in recommending distinct products based on common user intents they fulfil.
It involves identifying user intents and connecting them with products: a profile of user intents is extracted using user interactions (e.g. co-buy records, reviews) for each product listing. 
Then, a mapping from intents to product listings can be built to predict useful products based on user intents.

One significant challenge towards effective intent understanding is the vague definition of user intents, which precludes effective intent identification and can easily result in contaminated intent-product associations. In prior work \cite{yu_folkscope_2023,luo_alicoco2_2021}, user intents are often blended with ``product properties'' or ``similar products'', which we argue are related to the products and not the users. These shortcuts may benefit existing product recommendation benchmarks, but are not aligned with the intent understanding objective, namely, to retrieve superficially distinct kinds of products serving common intents~\cite{HMP2024}.  

Therefore, we propose a usage-centric paradigm for intent understanding (demonstrated in Figure \ref{fig:general_describe}). In this paradigm, user intents are focused on natural language predicative phrases, i.e. how users \textbf{use} a product;
also, instead of individual product listings, we aim to predict \textit{kinds of products} \textbf{useful} for an intent.
In particular, we define user intents as activities to accomplish (e.g. \textcolor{olive}{outdoor barbecue}) or situations to resolve (e.g. \textcolor{olive}{lower-back pain}); and,
kinds of products as clusters of product listings possessing the same category (e.g. \textcolor{purple}{scrub brush}) and property (e.g. \textcolor{purple}{stiff bristle}). Predicting at the level of the kinds of products guarantees that the list of relevant predictions is not endless.
Our task is a natural language reasoning task, closely related to commonsense reasoning \cite{sap_atomic_2019,bosselut_comet_2019}: ``The user has intent $I$'' entails ``The kind of product $P$ is useful for the user.''


Knowledge Graphs (KGs)
are important to many enterprises today, providing factual knowledge and structured data that steer many products and make them  more ready to be used in automatic processes and thus supporting more intelligent applications.
In this paper, we present an analysis of a SOTA E-Commerce intent knowledge graph, FolkScope \cite{yu_folkscope_2023}, which reported promising results on an intrinsic co-buy prediction task. Refactoring their KG to build associations between kinds of products and their usage user intents, we discover two unsatisfactory characteristics in their KG topology: 1) \textit{property-ambiguity}: generated user intents are poorly aligned with relevant product properties, such that 
the KG often maps user intents to kinds of products with relevant category but fairly random properties; 2) \textit{category-rigidity}: each intent is strongly associated with a single category of product, such that the KG is unable to recommend diverse products across different categories that serve common intents.

In light of these findings, we develop a
Product Recovery Benchmark, including an 
evaluation framework that aligns with the usage-centric paradigm, isolating product-specific confounders, such as product price or ratings. Also, we provide a dataset based on 
the Amazon Reviews Dataset (ARD) \cite{ni_justifying_2019} where we further validate the impact of the weaknesses in FolkScope.
All intent understanding methods developed on the ARD can be evaluated using this benchmark.

To summarize, in this paper: 1) we propose a usage-centric paradigm for intent understanding;
2) we introduce a product recovery benchmark featuring a novel evaluation framework, and report results with SOTA baselines; 3) we identify crucial weaknesses in existing SOTA as \textit{category-rigidity} and \textit{property-ambiguity}, and propose intent mining from user reviews as a promising future direction to address these issues.

\section{Usage-Centric Intent Understanding}
\label{sec:intents2kinds}

We propose a
usage-centric paradigm of intent understanding, focusing on usage user intents and the kinds of useful products, where the goal is to ground usage user intents in kinds of useful products. 
Differently from the ``informal queries'' in~\citet{luo_alicoco2_2021}, and similarly to~\citet{ding_mining_2015}, our usage user intents are generic eventualities/situations, independent of product ontologies. 




We introduce \textit{kinds of products} as the target granularity level, as it abstracts away the nuanced differences among individual listings, and yields a purely natural language setup, independent of product ontologies. 
It contains just enough information (category + property) to represent the product listings inside for intent understanding.

User intents rarely require combinations of properties in a product category. Therefore, to avoid generating factorial numbers of 
kinds of product, we impose a mild constraint that only one property is specified for each 
kind of product.

We demonstrate the specificity trade-off with an example below: for \textcolor{olive}{outdoor barbecues}, a \textcolor{purple}{stiff-bristle scrub brush} is useful for cleaning the grease on the grill. To that end, there are many listings of hard-bristle scrubs but the exact choice among them is irrelevant to the user intent and could be identified by downstream recommendation systems using other factors (customer habit, geo-location, etc.). However, the \textcolor{purple}{stiff bristle} property is essential for a listing to be suitable for \textcolor{olive}{outdoor barbecues}.
In short, grouping based on \textit{kinds of products} strikes a balance between sparsity that comes with specificity, and ambiguity that comes with generality.
 


\section{FolkScope Analysis}
\label{sec: folkscope_analysis}
\subsection{KG Refactoring}
\label{sec: folkdscope_kg}
We refactor FolkScope
based on our usage-centric intent understanding paradigm.
FolkScope KG connects products with their user intents, which are generated with OPT-30B \cite{zhang2022opt} when given pairs of co-bought products sourced from ARD \cite{ni_justifying_2019}, along with manually defined commonsense relations.



Among their 18 commonsense relations, we filter out all ``item'' relations as well as 3 ``function'' relations (\textit{SymbolOf}, \textit{MannerOf}, and \textit{DefinedAs}), since they are nominal in nature, and are irrelevant to product usage. We keep the remaining 5 predicative relations, \textit{UsedFor}, \textit{CapableOf}, \textit{Result}, \textit{Cause}, \textit{CauseDesire}, as legitimate user intents. 

To group the product listings into kinds of products, we take the fine-grained product categories from ARD (e.g. \textit{Kids' Backpacks}), 
and borrow the attributes under the relation \textit{PropertyOf} in the original FolkScope KG as properties.\footnote{These attributes 
do not fit the criteria for usage user intents, but they are 
acquired through generic LLM prompted summarization, and thus are borrowed 
as
product properties.}

We compute the association strengths from selected
user intents to 
common kinds of products by aggregation. 
Let $e(I_i, P_j)$ be the connection of intent $I_i$ with product listing $P_j$, $P_j$ belongs to a kind of products $K_k$. The association strength for edges in the refactored KG are then computed as: $e'(I_i, K_k) = \sum_{P_{j'} \in K_k} pmi(P_j, K_k)*e(I_i, P_j)$.
\footnote{The $pmi$ term penalizes product listings with multiple kinds of products (e.g. multiple properties in one listing).}

\subsection{Statistical Analysis}
\label{sec: folkscope:statistic}
We identify two major weaknesses of FolkScope KG under the usage-centric paradigm: it is over-specific about categories of useful products, but under-specific about the required properties of these products within each category. Intents in FolkScope tend to be associated with products having vague properties from few categories, rather than specific kinds of products across a variety of categories. 


\begin{figure}
    \centering
    \includegraphics[width=0.95\columnwidth]{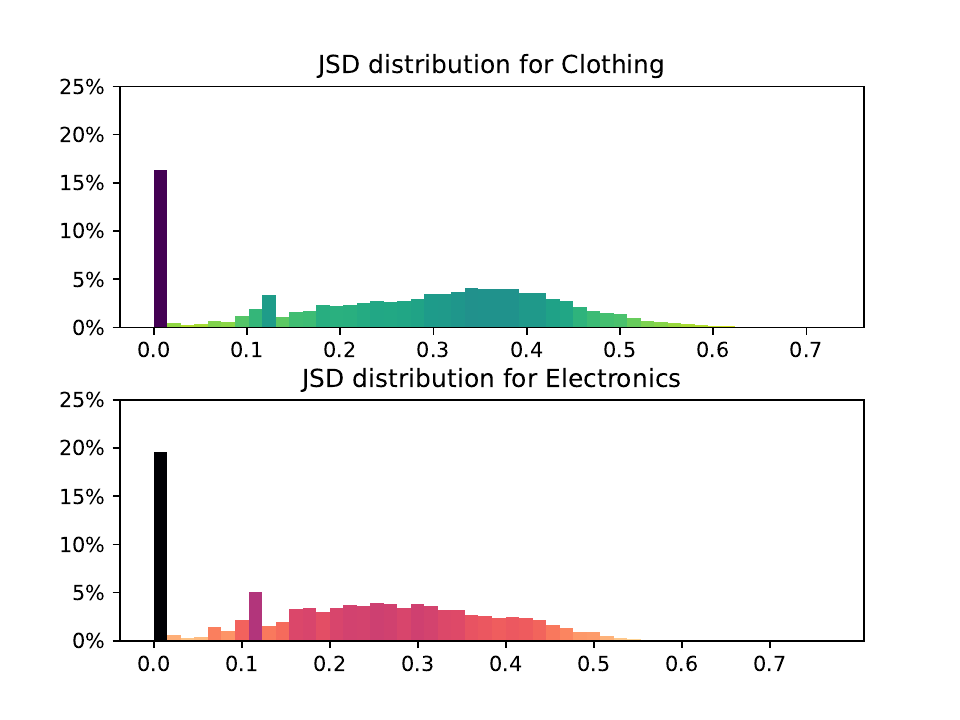}
    \caption{Histograms of Jensen-Shannon Divergence for each intent-category pair. Values are packed around 0: property-distributions of edge weights conditioned on intents are close to unconditioned frequency priors.}
    \label{fig:jsd_distribution}
\end{figure}

\paragraph{Property-Ambiguity}
For each 
user intent, we look into the distribution of its edge weights among 
kinds of products from one category with different properties. 
We compare these posterior edge-weight distributions, conditioned on intent, with the prior distributions across differently-propertied kinds of products within that category.
We calculate Jensen-Shannon Divergence (JSD)
between these conditional and prior distributions (see Figure~\ref{fig:jsd_distribution}): for up to 20\% 
of cases, JSD is $<$ 0.1, where only 2\% of cases have JSD $>$ 0.5. 

This shows, the KG's edge weights among differently-propertied kinds of products within the same category
are strongly predicted by their prior distribution, and are insensitive to the specific usages depicted by user intents. 
For example, for 
the user intent of \textcolor{olive}{outdoor barbecues}, 
its edge weights distribution among different kinds of \textcolor{purple}{scrub brush} products should depend on this specific usage scenario. In this case, a \textcolor{purple}{stiff bristle scrub brush} may receive much higher weights than other kinds of scrub brushes, rather than having the distribution align more closely with the prior distribution of kinds of \textcolor{purple}{scrub brush} products. We credit this to the mismatch between property and intent mining: each product listing may have multiple properties and serve multiple intents, but the mappings between these properties and intents are underspecified. 
\begin{figure}
    \centering
    \includegraphics[width=0.97\columnwidth]{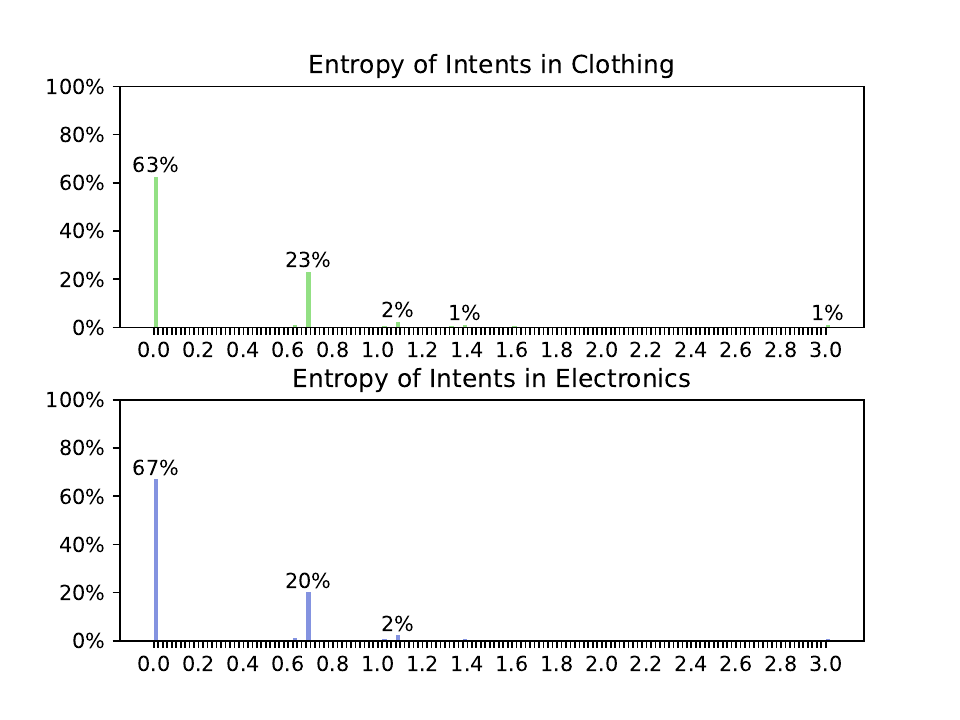}
    \caption{Histograms of category-entropy for each user intent. Values are concentrated at 0.0 and 0.7, meaning the intent is associated with only 1 / 2 categories. }
    \label{fig:entropy}
\end{figure}
\paragraph{Category-Rigidity}
In the refactored KG, we calculate the category diversity by measuring how diverse the edge weights are w.r.t. categories for one user intent. For each user intent, we add up its edge weights to kinds of products grouped by product categories (e.g. edge weights to \textcolor{purple}{stiff bristle \textbf{scrub brush}} and \textcolor{purple}{\textbf{scrub brush} with wooden handle} are added together), and compute the entropy of the converted category distribution.

Figure\,\ref{fig:entropy} shows the entropy meta-distributions: entropy values are concentrated in 2 narrow ranges, $[0, 0.02)$ and $[0.68, 0.70)$. We notice that an entropy in $[0, 0.02)$ indicates that the associations about this intent are focused on only one product category; $[0.68, 0.70)$ indicates that the associations are focused on two product categories. Therefore, from Figure \ref{fig:entropy} we can conclude that over 80\% of the intents are associated with only one or two categories. This category-rigidity in FolkScope hampers its ability to recommend diverse kinds of products, as we will discuss in \S\ref{sec:folkscope:results}.


\section{The Product Recovery Benchmark}
\label{sec:product_recovery}




\subsection{Benchmark Design}
\label{sec:productbench}
Following our intent understanding paradigm in \S\ref{sec:intents2kinds}, we introduce a usage-centric evaluation framework, which aims to recover kinds of products based on retrieved user intents.
Under this framework, an intent understanding method first predicts a profile of user intents for a product listing (using product description, user reviews, etc.). Then, using \textbf{solely} the predicted intent as input, the method recovers useful kinds of products based on its knowledge of E-Commerce demands (e.g. in symbolic KGs or LLMs).
The predictions are compared against: 1) \textit{bought-product-recovery}: kinds of product to which the current product belongs; 2) \textit{co-bought-product-recovery}: kinds of co-bought products that belong to other categories.

We take \textit{bought-product-recovery} as our main evaluation setup, since it focuses on intent-to-kinds-of-product associations. 
We also include the co-bought-product-recovery setup
to validate statistical findings on cross-category recommendation performance.
Compared to the product recommendation evaluation in \citet{yu_folkscope_2023}, this framework marginalizes factors inciting co-buy behaviour (e.g. brand loyalty, geolocation, etc.).

We instantiate the proposed evaluation framework with a product recovery benchmark, based on the ARD
\cite{ni_justifying_2019}, using available resources. We utilise the pool of product listings in ARD, enriched with product descriptions, category information, anonymized user purchase records and reviews. We additionally borrow kinds of products from refactored FolkScope, as in \S\ref{sec: folkdscope_kg}.\footnote{Our elicitation procedure is corpus-agnostic, we empirically select ARD as it is the largest available dataset; we acknowledge that re-using information from FolkScope may grant it an unfair advantage, however, we show below, that it nevertheless suffers from the aforementioned weaknesses and fails to perform intent understanding effectively.}

\paragraph{Evaluation metric}
Following prior work \cite{CHEN201344}, we measure success by Mean Reciprocal Rank (MRR) of gold kinds of products in the predicted distributions as shown in Eq. \ref{eq:mrr_max}.
In case multiple gold kinds of products are assigned for a product listing, 
we calculate the $\text{MRR}_{\max}$ using the highest-ranking hit. 

\begin{equation}
    \text{RR}_{\max}(l) = \max_{c \in C_{\text{gold}}\left ( l \right )}\left (\text{rank}(c)^{-1} \right )
\end{equation}
\begin{equation}
    \text{MRR}_{\max} = \frac{\sum_{l\in L} \text{RR}_{\max}(l)}{|L|}
\label{eq:mrr_max}
\end{equation}
where RR represents the Reciprocal Rank, $C_{\text{gold}}(l)$ are the gold clusters for the listing $l$ and $L$ is the set of all listings in the benchmark.

\subsection{Experiments and Results}
\label{sec:folkscope:results}

\begin{table}[t]
\centering
\begin{tabular}{lcc}
\toprule
Models            & Clothing     & Electronics        \\ 
\midrule
FolkScope & \textcolor{blue}{0.192} & \textcolor{teal}{0.263} \\
FolkScope $-$ properties & \textcolor{blue}{0.116} & \textcolor{teal}{0.166} \\
\midrule
FolkScope $+$ GPT & \textcolor{blue}{0.187} & \textcolor{teal}{0.257} \\
\bottomrule
\end{tabular}

\caption{\label{tab: test_mrr}$\text{MRR}_{\max}$ for \textit{bought-product-recovery} task.}
\end{table}

We evaluate the FolkScope KG (refactored in \S\ref{sec: folkdscope_kg}) with the Product Recovery benchmark. We offer the baseline results in Table \ref{tab: test_mrr}, and highlight below the impact of weaknesses discussed in \S\ref{sec: folkscope:statistic}.

\paragraph{Property-Ambiguity}
To understand how property ambiguity affects FolkScope performance, we compare it with another prior property baseline derived from it: for each evaluation entry, we corrupt the FolkScope predictions by replacing the property in the predicted kinds of products based on the property popularity. The popularity of a property is defined as the frequency with which it appears in the product listings that belong to the same fine-grained category (e.g. \textcolor{purple}{scrub brush}) as the evaluation entry (kinds of products). To avoid making duplicate predictions after substitution, if multiple kinds of products from the same category are predicted, we draw properties top-down w.r.t. popularity for each prediction.

From Table \ref{tab: test_mrr}, we observe that \textit{FolkScope $-$ properties} reached respectable performance with only moderate regression from FolkScope predictions. This limited MRR gap shows the impact of property-ambiguity, where performance gains could be expected with better property alignment.

\paragraph{Category-Rigidity}
To validate the category-rigidity observation in \S\ref{sec: folkscope:statistic}, we also evaluate the FolkScope KG in the co-bought-product-recovery setup, where we specifically use it to predict kinds of co-bought products in \textbf{other categories}.

In this setup, we observe low $\text{MRR}_{\max}$ of \textcolor{blue}{0.077} and \textcolor{teal}{0.033} for \textit{Clothing} and \textit{Electronics} domains, respectively:
the FolkScope KG cannot effectively recommend superficially distinct kinds of products connected with the same user intents.

Notably, between the two domains, FolkScope reaches a slightly higher $\text{MRR}_{\max}$ in \textit{Clothing}. This is consistent with our findings in Figure \ref{fig:entropy}, where category-entropy values are slightly more spread than in \textit{Electronics} (i.e. category rigidity is less severe).

\paragraph{LLM Rerank}
We also evaluate LLM performance on usage-centric intent understanding using our benchmark, using GPT-3.5-turbo \cite{brown_language_2020}. Ideally, we would like the LLM to predict useful kinds of products end-to-end. However, due to the difficulty of reliably matching LLM predictions with gold kinds of products\footnote{In Appendix \ref{appendix:gpt_eval}, we include an LLM-only baseline using GPT-4 as matching metric, where we find it underperforming FolkScope baseline, and find GPT-4 metric over permissive.}, we instead adopt a re-ranking paradigm, where we prompt the LLM to re-rank the top-10 kinds of products predicted by FolkScope.

As Table \ref{tab: test_mrr} shows, we observe no clear benefit with LLM-reranking.
We investigate this failure by looking into where hits are met in the predictions. From Table \ref{statistical_analyses:hit}, we find that most hits are either at first or not in the top 10.
These polarized distributions leave little room for re-ranking to take effect.

We raise the warning that dataset artefacts from the common source corpus (AWD) could be behind this abnormally high hit-at-1 rate (compared with the $\text{MRR}_{\max}$ value), where the reported $\text{MRR}_{\max}$ values may have been inflated. Due to the lack of another large E-Commerce Reviews corpus, we leave further investigations for future work.

\section{Discussions and Conclusion}

In this paper, we revisit intent understanding from a usage-centric perspective, as a natural language reasoning task, to detect superficially distinct kinds of products useful for common usage intents. We
developed a Product Recovery benchmark, and
investigated two weaknesses of the SOTA FolkScope KG in supporting usage-centric intent understanding: \textit{Property Ambiguity} and \textit{Category-Rigidity}.

 \begin{table}[t]
 \centering
\begin{tabular}{ccc}
\toprule
               & Clothing & Electronics \\ \midrule
hit@1     & 16\%     & 22\%        \\ \midrule
hit > 10 & 73\%     & 63\%        \\ \bottomrule
\end{tabular}
\caption{\label{statistical_analyses:hit}The ratio of hit being the first in the prediction list and not in the top-10 of the prediction list.}
\end{table}

We advocate for adopting the usage-centric intent understanding paradigm, and for 
considering user reviews, in addition to co-buy records.
Desired product properties and their respective intents are likely to co-occur in product reviews, relieving property-ambiguity; the same usage intents tend to be described consistently in user reviews across different categories, relieving category-rigidity.

As for future work, one idea is to use our proposed benchmarks to test some entailment graphs in E-commerce. We might further investigate some abstract inference capabilities that are related to conceptual understanding.
\section*{Limitations}
In this paper, we have proposed to study E-Commerce intent understanding from a usage-centric perspective. Due to the lack of consistent task definition and limited computational budget, we are only able to analyse one SOTA intent understanding KG (namely FolkScope) and one SOTA LLM. We encourage more research attention on the usage-centric E-commerce intent understanding task for a more diverse landscape.

We have established that weaknesses of Property Ambiguity and Category Rigidity exist in the SOTA KG, and we have offered a principled hypothesis that utilizing genuine user reviews could help with these weaknesses. However, due to limits to the scope of this paper, we do not provide empirical evidence for this hypothesis and leave it as a promising direction of future work.

We note that as this paper is related to recommendation, there exists risks that methods developed on the Product Recovery Benchmark may be used to bias customer decisions; on the other hand, we also note that our task definition is purely natural language and does not involve any individual product listings, therefore it would not bias customer choices among directly competing listings of the same kinds of products.

\section*{Acknowledgements}

We would like to thank the reviewers for their valuable comments and suggestions. This work was partly funded by a Mozilla PhD scholarship at Informatics Graduate School and by the University of Edinburgh Huawei Laboratory.

\bibliography{references,references_manual}
\bibliographystyle{acl_natbib}
\appendix

\label{sec:appendix}
\section{Implementation Details}
\subsection{Benchmark data split}
We follow \citet{yu_folkscope_2023}, and we split product instance in FolkScope KG into training, validation and test splits with respective portions of 80\%, 10\% and 10\%. Please refer to Table~\ref{tab:stats} for detailed statistics. Note that \textit{Clothing} stands for the ``Clothing, Shoes and Jewelry'' domain in the Amazon Reviews Dataset, and \textit{Electronics} simply stands for the ``Electronics'' domain in the Amazon Reviews Dataset.


\begin{table}[ht]
\begin{tabular}{L{2.5cm}ccc}
\toprule
\textbf{Categories}        & \textbf{Train} & \textbf{Validation} & \textbf{Test}  \\
\toprule
Clothing & 30296 &  2027       & 2088  \\ 
\midrule
Electronics                & 85086 &   7853    & 7900 \\ 
\bottomrule
\end{tabular}
\caption{\label{t1} Number of product listings in the training, validation and test set. Please note that we drop product listings that lack related kinds of products, so the ratio of the number of instances across the splits are not exactly equal to 8:1:1.}
\label{tab:stats}
\end{table}


\subsection{GPT-3.5-turbo Re-ranking}
\label{appendix: gpt-reranking}
For each product listing $l$, when there is no predicted kind of products given a set of related user intents, we mark the $\text{RR}_{\max}(l)$ as 0 both before and after re-ranking.
\subsubsection{Re-ranking Prompt}


\begin{displayquote}
A product is suitable for the following purposes:\\
\{\textit{Intents}\}\\

\noindent Please rank the following categories in order of likelihood that the product belongs to them (most likely to least likely): \\
\{\textit{kinds of products list}\}
...\\
Answer:\\
1.\\
\noindent
\end{displayquote}

We fill \textit{Intents} with a set of mined user intents and \textit{kinds of products list} with the top 10 predictions for kinds of products.

Note that in this setting and in \S~\ref{appendix:gpt:prompt:prediction}, we still use the term ``category'' in LLM prompts to refer to kinds of products, because during preliminary experiments we found that LLMs do not respond well to the term ``kind of product''.

\section{GPT End-to-End Evaluation}
\label{appendix:gpt_eval}
We perform an additional experiment to directly predict kinds of products in an end-to-end setup, with an LLM, for our proposed product recovery task. Again, we use GPT-3.5-turbo as the LLM and design the zero-shot prompt as in \S\ref{appendix:gpt:prompt:prediction}. However, due to the absence of the complete ontology of the Amazon Reviews Dataset, it is challenging for GPT-3.5-turbo to predict the exact ground truth kinds of products. To sidestep the difficulty of evaluating whether the predicted strings are semantically identical to the ground truth labels, we use GPT-4 to judge whether there is a match between predicted and ground truth labels. The relevant prompt is specified in \S\ref{appendix:gpt:prompt:evaluation}.
The detailed evaluation results is presented in Table \ref{tab: gpt4-eval}.

\begin{table}[t]
\begin{center}
\begin{tabular}{lcc}
\toprule
 & Clothing & Electronics \\
 \midrule
GPT-3.5-turbo & 0.511   & 0.543     \\ 
\midrule
FolkScope          & 0.527   & 0.671     \\ \bottomrule
\end{tabular}
\end{center}

\caption{$\text{MRR}_{\max}$ score when evaluating using GPT-4 as the judge for matching. Values for GPT-3.5-turbo and our baseline refactored FolkScope KG are both higher in absolute values due to the more benign matching criterion; the LLM baseline with GPT-3.5-turbo does not outperform the KG baseline.}
\label{tab: gpt4-eval}
\end{table}
From Table \ref{tab: gpt4-eval}, we can observe that GPT-3.5-turbo does not outperform the FolkScope KG baseline
on the product recovery benchmark. Compared to the strict string matching results in Table \ref{tab: test_mrr}, GPT-4 evaluation has a significantly more permissive criterion on matching, yielding much higher $\text{MRR}_{\max}$ values. We find many of these ``matched'' verdicts by GPT-4 to be spurious (see Table \ref{tab:gpt-4:wrong}), and conclude that GPT-4 cannot easily achieve reliable matching for the product recovery benchmark, and more robust criteria are needed before replacing the exact match criterion. 
\begin{table}[t]
\begin{center}
\begin{tabular}{lll}
\toprule
Experiment        & Clothing & Electronics \\ \midrule
LLM Rerank    & 3.86 \$  & 1.38 \$    \\ \midrule
LLM End-to-End & 15.57 \$ & 14.56\$    \\ \bottomrule
\end{tabular}
\end{center}
\caption{API costs of our LLM-related experiments. For the LLM Rerank experiment, we re-rank all the data samples in the test set while for the End-to-End evaluation, we only sample 1000 data samples in the test set. \label{tab:llm_cost} }
\end{table}

\begin{table*}[htbp]
\begin{tabular}{l}
\toprule
\textbf{Ground truth kinds of products}                                                             \\\midrule
1. Clothing, Shoes \& Jewelry|Costumes   \& Accessories|Men|Accessories \#\#\# Wandering Gunman \\
2. Clothing, Shoes \& Jewelry|Costumes   \& Accessories|Men|Accessories \#\#\# Holster          \\
3. Clothing, Shoes \& Jewelry|Costumes   \& Accessories|Men|Accessories \#\#\# Western          \\ \midrule
\textbf{GPT-3.5-turbo prediction}                                                                       \\ \midrule
1. Clothing, Shoes \&   Jewelry|Men|Costumes|Western \#\#\# authentic                            \\
…                                                                                               \\ \bottomrule
\toprule
\textbf{Ground truth kinds of products}                                            \\ \midrule
1. Clothing, Shoes \&   Jewelry|Women|Jewelry|Earrings|Stud \#\#\# Jewelry  \\
2. Clothing, Shoes \&   Jewelry|Women|Jewelry|Earrings|Stud \#\#\# Gemstone \\
3. Clothing, Shoes \&   Jewelry|Women|Jewelry|Earrings|Stud \#\#\# Sterling Silver        \\ \midrule
\textbf{GPT-3.5-turbo prediction}                                                    \\ \midrule
1. Clothing, Shoes \&   Jewelry|Women|Earrings|Stud Earrings \#\#\# elegant and beautiful \\
…                                                                           \\ \midrule
\end{tabular}
\caption{Here we list two examples that GPT-4 validate with $\text{RR}_{\max} = 1$. In the first example, it validates the first prediction as true by matching the ``property'' part of the ground truth 3 with the main category of prediction 1. In the second example, the ``property'' part of prediction 1 is too general compared to all the ground truth kinds of products, but it still validates it as true.}
\label{tab:gpt-4:wrong}
\end{table*}
\subsection{Prompt Examples}

\subsubsection{Kinds of Products Prediction}
\label{appendix:gpt:prompt:prediction}

\begin{displayquote}
Intents: \\
\{\textit{intents}\}\\
Given the intents, please predict the top 10 kinds of products that will be useful for these intents. \\
A kind of product is the concatenation of a fine-grained category from the Amazon Review Dataset and a useful property. For example: Clothing, Shoes \& Jewelry|Men|Watches|Wrist Watches \#\#\# leather. \\
Kinds of products:\\
1.
\noindent
\end{displayquote}

\subsubsection{Prediction Evaluation}
\label{appendix:gpt:prompt:evaluation}

\begin{displayquote}
Here is a list of predicted categories:\\
\{\textit{prediction}\}\\
Validate each prediction based on the ground truth categories[T/F].\\
Each prediction can be considered true when it is similar to one of the ground truth categories.\\
Ground truth categories:\\
\{\textit{ground truth}\}
\end{displayquote}

\section{Computational Budget}

\subsection{Main Experiments}
\label{appendix: exp: main}
All the benchmark construction and evaluation has been performed using 2 x Intel(R) Xeon(R) Gold 6254 CPUs @ 3.10GHz. 
\paragraph{FolkScope KG Refactoring} We converted all the intents generated by FolkScope without applying any of its proposed filters based on the graph evaluation results on the validation set. The whole graph generation for both domains takes around 24 hours in total.
\paragraph{FolkScope Intents Evaluation} We need around 71 and 6 hours for evaluating the intents for the test set of the Clothing and Electronics domain respectively.

\subsection{LLM Experiments}
We mainly use GPT-3.5-turbo and GPT-4 for our LLM-related experiments. Please refer to Table~\ref{tab:llm_cost} for details about the relevant costs. For both models, we keep the default query parameters from OpenAI, and set the temperature to 0 to promise reproducability.
\label{appendix: exp: LLM}

\section{Artifact Licenses}
\label{appendix:licensing}
\textbf{Amazon Reviews Dataset}: Limited license for academic research purposes and for non-commercial use (subject to \href{https://www.amazon.com/gp/help/customer/display.html/ref=footer_cou?ie=UTF8&nodeId=508088}{Amazon.com Conditions of Use})\\
\textbf{FolkScope}: MIT license

\end{document}